\documentclass{article}

\usepackage[final]{neurips_2020}

\usepackage[utf8]{inputenc} 
\usepackage[T1]{fontenc}    
\usepackage{hyperref}       
\usepackage{url}            
\usepackage{booktabs}       
\usepackage{amsfonts}       
\usepackage{nicefrac}       
\usepackage{microtype}      

\title{The Challenge of Diacritics in Yor\`{u}b\'{a} Embeddings}

\author{%
  Tosin P. Adewumi\thanks{Corresponding author}
  , Foteini Liwicki \&
  Marcus Liwicki \\
  \\
  EISLAB, SRT\\
  Luleå University of Technology, Sweden\\
  \texttt{firstname.lastname@ltu.se} \\
}

\begin{document}

\maketitle

\begin{abstract}
The major contributions of this work include the empirical establishment of a better performance for Yor\`{u}b\'{a} embeddings from undiacritized (normalized) dataset and provision of new analogy sets for evaluation.
The Yor\`{u}b\'{a} language, being a tonal language, utilizes diacritics (tonal marks) in written form.
We show that this affects embedding performance by creating embeddings from exactly the same Wikipedia dataset but with the second one normalized to be undiacritized.
We further compare average intrinsic performance with two other work (using analogy test set \& WordSim) and we obtain the best performance in WordSim and corresponding Spearman correlation.
\end{abstract}

\section{Introduction}
The Yor\`{u}b\'{a} language is spoken by about 40 million people in West Africa and around the world \citep{fakinlede2005beginner}.
Of the various dialects around, the standard Yor\`{u}b\'{a} language (pioneered by Bishop Ajayi Crowther) is the focus of this paper.
Standard Yor\`{u}b\'{a} orthography uses largely the Latin alphabet and is the widely spoken dialect among the educated \citep{bamgbose2000grammar}.
Yor\`{u}b\'{a} has 25 letters in its alphabet,
though counting the 5 nasal vowels makes it 30 \citep{fakinlede2005beginner, asahiah2017restoring}.
Being a tonal language, 3 diacritics are used on vowels based on syllables per word: depression tone (grave), optional mid tone and elevation tone (acute) \citep{church1913dictionary}.
Besides these differences between the English and the Yor\`{u}b\'{a} languages,
Yor\`{u}b\'{a} has no gender identification for verbs or pronouns \citep{nurse2010verbal}. 
Yor\`{u}b\'{a} verb tenses are usually determined within context and remain mostly the same in spelling and tone \citep{lamidi2010tense, uwaezuoke2017contrastive}.

The research question we address in this work is "Do diacritics affect the performance of Yor\`{u}b\'{a} embeddings and in what way?"
This is because it has been observed by
\citet{asubiaro2014effects} that web-search without diacritics produced more relevant results than search-words containing them, while evaluating four popular search engines.
He also found out that the effectiveness of two of the search engines were adversely affected with diacritics.
Thus, the objectives in this work include providing optimal Yor\`{u}b\'{a} embeddings and creating new analogy test set to evaluate the embeddings.
Optimal hyper-parameter combination for the embeddings were chosen based on the work by \citet{adewumi2020word2vec, adewumi2020exploring}.
The heavily pre-processed (cleaned) Wikipedia dataset and the new analogy test set will provide valuable contributions to the natural language processing (NLP) community for the Yor\`{u}b\'{a} language, a low-resource language.
The rest of this paper include the related work, the methodology, the results \& discussion and the conclusion sections.

\section{Related work}
Initial effort by Ajayi Crowther to document Yor\`{u}b\'{a} barely had tonal marks
\citep{bowen1858grammar}.
In fact, early dictionary by \cite{church1913dictionary} had minimal diacritics compared to the modern Yor\`{u}b\'{a} dictionary by \citet{pamela2005yoruba}.
This implies the language has been evolving and usage or discernment of diacritics between then and now is different.
Revised efforts, later, standardized the diacritics and afforded others the opportunity to expand the work \citep{asahiah2017restoring, fagborun1989disparities}.
For example, the word \textit{abandon} in the \citet{church1913dictionary} dictionary is \textit{k\d{o}-sil\d{e}} while it is \textit{k\d{\`{o}}-s\'{i}l\d{\`{e}}} in the modern Lexilogos dictionary\footnote{www.lexilogos.com/english/yoruba\_dictionary.htm} and that by \cite{pamela2005yoruba}.

Absence of diacritics made contextual semantics of words, probably, more important back then than they are today, given that some words with the same spelling can have different meanings, depending on the context.
Even the English language has words which are spelled the same way but pronounced differently and have different meanings (homographs), exposed by context, e.g. \textit{lead, row} or \textit{fair}.
Given the relative challenge of producing Yor\`{u}b\'{a} diacritics among some users,
the versions without diacritics or partial diacritics have been increasing \citep{asubiaro2014effects, asahiah2017restoring, fagborun1989disparities}.
This has led some to push for the normalization (restricting diacritized letters to their base versions) of the Yor\`{u}b\'{a} language, especially in electronic media \citep{asubiaro2014effects}.
This attempt may also lead to canonicalization of Yor\`{u}b\'{a} text, through the relationship between diacritized and undiacritized words that will be established.


Other researchers, like \citet{asahiah2017restoring} argue that diacritic restoration is a necessity.
However, their own research showed the possible challenge for beginners of adding diacritics when the corpus they utilized had roughly the same percentage for the 3 diacritic marks \citep{asahiah2017restoring}.
Yor\`{u}b\'{a} diacritic restoration is being undertaken by some researchers from word-level, syllable-level or character-level restoration and some of the methods for automatic diacritization utilize Machine Learning (ML) methods \citep{asahiah2017restoring}.

Word embeddings have shortcomings, such as displaying biases in the data they are trained on \citep{bolukbasi2016man}.
However, they can be very useful for practical NLP applications.
For example, subword representations have proven to be helpful when dealing with out-of-vocabulary (OOV) words and \citet{thomason2020jointly} used word  embeddings to guide the parsing of OOV words in their work on meaning representation for robots.
Intrinsic tests, in the form of word similarity or analogy tests, despite their weaknesses, have been shown to reveal meaningful relations among words in embeddings, given the relationship among words in context \citep{mikolov2013efficient, pennington2014glove}.
It is inappropriate to assume such intrinsic tests are sufficient in themselves, just as it is inappropriate to assume one particular extrinsic (downstream) test is sufficient to generalise the performance of embeddings on all NLP tasks \citep{gatt2018survey, faruqui2016problems, adewumi2020word2vec, adewumi2020corpora}.

\section{Methodology}
Three Yor\`{u}b\'{a} training datasets were used in this work.
They include the cleaned 2020 Yor\`{u}b\'{a} Wikipedia dump containing diacritics to different levels across articles \citep{yowiki2020}, a normalized (undiacritized) version of it and the largest, diacritized data used by \cite{alabi2020massive}.
The original Yor\`{u}b\'{a} Wikipedia dump has a lot of vulgar content, in addition to English, French \& other language content.
Manual cleaning brought the file size down to 182MB from 1.2GB, after using a Python script to remove much of the HTML tags, from the initial raw size of 1.7GB.
Using the recommended script by \cite{grave2018learning} to preprocess the original dataset did not work as intended, as it retained all the English \& foreign content and removed characters with diacritics from the Yor\`{u}b\'{a} parts.
An excerpt from the cleaned Wikipedia data, discussing about the planet Jupiter, is given below:
\begin{quote}
    Awo osan ati brown inu isujo J\'{u}p\'{i}t\'{e}r\`{i} wa lati iwusoke awon adapo ti won unyi awo w\d{o}n pada nigba ti w\d{o}n ba dojuko im\d{o}le [[ultraviolet]] lati \d{o}d\d{o} Orun. Ohun to wa ninu awon adap\d{o} w\d{o}nyi ko daju, botilejepe fosforu, sulfur tabi boya [[hydrocarbon|haidrokarbon]] ni w\d{o}n je gbigbagb\d{o} pe w\d{o}n je.
\end{quote}

The authors created two analogy test sets: one with diacritics and an exact copy without diacritics.
However, all results reported in the next section were for the standard diacritic versions of the analogy and WordSim sets.
The results based on the undiacritized WordSim set for both Wiki versions were poorer than what is reported in the next section but the undiacritized Wiki version still gave better results than the diacritized against that set.
Creating the analogy sets (containing over 4,000 samples each) was challenging for some of the sections in the original Google version by \cite{mikolov2013efficient}.
For example, in the \textit{capital-common-countries} sub-section of the semantic section, getting consistent representations of some countries, like \textit{Germany}, is difficult, as it is translated as \textit{J\d{e}mani} by some or \textit{Jamani} by others.
A very useful resource is Lexilogos, which translates from English to Yor\`{u}b\'{a} and, importantly, displays a number of contextual references where the translation is used in Yor\`{u}b\'{a} texts.
The analogy sets are smaller versions of the original, with 5 sub-sections in the semantic section and only 2 sub-sections in the syntactic section.
All datasetsa and relevant code used are available for reproducibility of these experiments.\footnote{https://github.com/tosingithub/ydesk}
Four samples from the \textit{gram2-opposite} of the diacritized version are given below:

\begin{quote}
w\'{a} l\d{o} \`{a}gb\`{a} \d{o}d\d{o}\\
w\'{a} l\d{o} \`{o}w\'{u}r\d{o} \`{i}r\d{o}l\d{e}\\
w\'{a} l\d{o} \d{o}t\'{a} \d{\`{o}}r\d{\'{e}} \\
w\'{a} l\d{o} nl\'{a} k\'{e}ker\'{e}
\end{quote}

Two types of embedding (word2vec and subword) per dataset were created, using the combination: skipgram-negative sampling with window size 4.
The minimum and maximum values for the character ngram are 3 and 6, respectivley, though the embedding by \citet{grave2018learning} used ngram size of 5.
Each embedding creation and evaluation was run twice to take an average, as reported in the next section.
A Python-gensim \citep{rehurek_lrec} program was used to conduct the evaluations after creating the embeddings with the original C++ implementation by \cite{grave2018learning}.
The Yor\`{u}b\'{a} WordSim by \cite{alabi2020massive} was also used for intrinsic evaluation.
This Yor\`{u}b\'{a} WordSim was based on the original English version by \cite{finkelstein2001placing}, containing a small set of 353 samples.
However, the Yor\`{u}b\'{a} version had a few issues, which we corrected before applying it.
For example, \textit{television} is translated as \textit{t\d{e}lif\'{o}s\`{i}\d{\`{o}}n\`{u}} instead of \textit{t\d{e}lif\'{i}\d{s}\d{\`{o}}n}, in one instance, and the bird \textit{crane} is translated as \textit{ot\'{i}-br\'{a}\'{n}d\`{i}} (brandy) instead of \textit{w\'{a}d\`{o}w\'{a}d\`{o}}, according to the Yor\`{u}b\'{a} dictionary.

\section{Results \& discussion}
Tables \ref{sample-table} \& \ref{sample-table2} show results from the experiments while table \ref{quality} gives nearest neighbor result for the random word \textit{iya (mother or affliction, depending on the context or diacritics)}.
Average results for embeddings from the 3 training datasets and the embedding by \citet{grave2018learning} are tabulated: Wiki, U\_Wiki, C3 \& CC, representing embeddings from the cleaned Wikipedia dump, its undiacritized (normalized) version, the diacritized data from \cite{alabi2020massive} and the Common Crawl embedding by \cite{grave2018learning}, respectively.
Performance of the original, contaminated Wikipedia dump was poorer than the cleaned version reported here, hence, it was left out from the table.
It can be observed from table \ref{sample-table} that the cleaned Wiki embedding have lower scores than the C3, despite the larger data size of the Wiki.
This may be attributed to the remaining noise in the Wiki dataset.
Inspite of this noise, the exact undiacritized version (U\_Wiki) outperforms C3, giving the best WordSim score \& corresponding Spearman correlation.
This seems to show diacritized data affects Yor\`{u}b\'{a} embeddings.
The negative effect of noise in the Wiki word2vec embedding seems to reduce in the subword version in table \ref{sample-table2}.

The best analogy score is given by the embedding from \cite{grave2018learning}, though very small.
The performance of the embeddings are much lower for analogy evaluations than their English counterparts as demonstrated by \citet{adewumi2020word2vec}, though the comparison is not entirely justified, since different dataset sizes are involved.
Other non-English work, however, show it's not unusual to get lower scores, depending, partly, on the idiosyncrasies of the languages involved \citep{adewumi2020exploring, koper2015multilingual}.
NLP downstream tasks, such as named entity recognition (NER), with significance tests, will be the definitive measure for the performance of these embeddings, and this is being considered for future work.



\begin{table}
  \caption{Yor\`{u}b\'{a} word2vec embeddings intrinsic scores (\%)}
  \label{sample-table}
  \centering
  \begin{tabular}{lrlll}
    \toprule
    Data & Vocab & Analogy & WordSim & Spearman\\
    \midrule
    Wiki & 275,356 & 0.65 & 26.0 & 24.36\\
    U\_Wiki & 269,915 & \textbf{0.8} & \textbf{86.79} & \textbf{90} \\
    C3 & 31,412 & 0.73 & 37.77 & 37.83\\
    \bottomrule
  \end{tabular}
\end{table}

\begin{table}
  \caption{Yor\`{u}b\'{a} subword embeddings intrinsic scores (\%)}
  \label{sample-table2}
  \centering
  \begin{tabular}{lrlll}
    \toprule
    Data & Vocab & Analogy & WordSim & Spearman \\
    \midrule
    Wiki & 275,356 & 0 & 45.95 & 44.79 \\
    U\_Wiki & 269,915 & 0 & 72.65 & 60 \\
    C3 & 31,412 & 0.18 & 39.26 & 38.69 \\
    CC & 151,125 & 4.87 & 16.02 & 9.66 \\
    \bottomrule
  \end{tabular}
\end{table}

\begin{table}[hbt!]
\caption{Example qualitative assessment of undiacritized word2vec model}
\label{quality}
\centering
\begin{tabular}{c|c}
\textbf{\footnotesize{Nearest Neighbor}} & \textbf{\footnotesize{Result}} \\
\hline
\footnotesize{iya} & \footnotesize{AgnEs (0.693), Arnauld (0.6798), ol\d{o}lajul\d{o} (0.678), Rabiatu (0.6249), Alhaja (0.6186),..} \\
\hline
\end{tabular}
\end{table}

\section{Conclusion}
The Yor\`{u}b\'{a} language is a tonal language and performance in NLP is affected, depending on diacritics, as shown in this work.
It appears it is advantageous normalizing diacritized texts before working on them for NLP purposes, as they produce better intrinsic performance, generally.
Our embeddings, based on normalized text, achieved better instrinsic performance than others tested.
Future work will involve utilizing the embeddings in downstream tasks, such as NER, using state-of-the-art (SotA) architectures.
Such downstream tasks will serve as the definitive measure for evaluating these embeddings.
There's ongoing effort on the sizable NER dataset to achieve this.



\section*{Broader Impact}
The broader impact of this paper is the insight it provides for NLP researchers in Yor\`{u}b\'{a} language with regards to the differences in performance, based on diacritics.
It provides 2 new analogy test sets for evaluating Yor\`{u}b\'{a} embeddings, depending on diacritics or the lack of it, and also provides an improved WordSim set.
Furthermore, a heavily preprocessed Wikipedia dataset for training embeddings is provided, in the diacritized and undiacritized versions.


\begin{ack}
The work in this project is partially funded by Vinnova under the project number 2019-02996 "Språkmodeller för svenska myndigheter".
\end{ack}

\bibliography{ref}
\bibliographystyle{apalike}

\end{document}